\PassOptionsToPackage{svgnames,dvipsnames,table}{xcolor}
\documentclass[sigconf,natbib=true,anonymous=false]{acmart}
\AtBeginDocument{%
  \providecommand\BibTeX{{%
    \normalfont B\kern-0.5em{\scshape i\kern-0.25em b}\kern-0.8em\TeX}}}

\setcopyright{acmlicensed}
\copyrightyear{2024}
\acmYear{2024}
\acmDOI{XXXXXXX.XXXXXXX}

\acmConference[DCAI24]{The 3rd International Workshop on Data-Centric AI at CIKM 2024.}{October 25, 2024}{Boise, ID}
\acmISBN{978-1-4503-XXXX-X/18/06}

\usepackage{times}
\usepackage{latexsym}

\usepackage[T1]{fontenc}
\usepackage[utf8]{inputenc}

\usepackage{microtype}

\usepackage{adjustbox}
\usepackage{booktabs}
\usepackage{graphicx}
\usepackage{lipsum}
\usepackage{xspace}
\usepackage{multicol}
\usepackage{mathrsfs}
\usepackage{multirow}
\usepackage{caption}
\usepackage{enumitem}
\usepackage{changepage}
\usepackage{tabularx}
\usepackage{comment}
\usepackage{subcaption}
\usepackage{pdflscape}
\usepackage{textcomp}
\usepackage{url}
\usepackage{amsfonts}
\usepackage{bbm}
\usepackage{arydshln}
\usepackage{makecell}
\usepackage{stfloats} % for positioning of figure* or table* on the same page

\usepackage{microtype}
\usepackage{array}
\newcolumntype{C}[1]{>{\centering}m{#1}}
\newcommand\BibTeX{B\textsc{ib}\TeX}

\newcommand{\augtrv}{\textsc{AugTriever}\xspace}
\newcommand{\contrv}{\textsc{Contriever}\xspace}
\newcommand{\bm}{\textsc{Bm25}\xspace}
\newcommand{\spider}{\textsc{Spider}\xspace}
\newcommand{\spar}{\textsc{Spar~\(\Lambda\)}\xspace}
\newcommand{\cpt}{\textsc{CPT}\xspace}

\newcolumntype{R}[1]{>{\raggedleft\arraybackslash}p{#1}}
\newcolumntype{L}[1]{>{\raggedright\arraybackslash}p{#1}}
\newcolumntype{C}[1]{>{\centering\let\newline\\\arraybackslash\hspace{0pt}}m{#1}}

\newcommand{\bestcell}{\cellcolor{BrickRed!30}}
\newcommand{\secbestcell}{\cellcolor{RoyalBlue!20}}

\usepackage{color, colortbl}
\definecolor{Gray}{gray}{0.90}
\definecolor{White}{gray}{1.0}
\definecolor{LightCyan}{rgb}{0.88,1,1}

\newcolumntype{g}{>{\columncolor{Gray}}c}
\newcolumntype{w}{>{\columncolor{White}}c}

\begin{document}

%%
%% The "title" command has an optional parameter,
%% allowing the author to define a "short title" to be used in page headers.
\title{AugTriever: Unsupervised Dense Retrieval and Domain Adaptation by Scalable Data Augmentation}

%%
%% The "author" command and its associated commands are used to define
%% the authors and their affiliations.
%% Of note is the shared affiliation of the first two authors, and the
%% "authornote" and "authornotemark" commands
%% used to denote shared contribution to the research.
\author{Rui Meng*}
\author{Lifu Tu}
\author{Meghana Bhat}
\affiliation{%
    \country{}
}

\author{Ye Liu}
\author{Ning Yu}
\author{Yingbo Zhou}
\email{ruimeng@salesforce.com*}
\affiliation{%
  \institution{Salesforce Research}
  \country{}
}

\author{Semih Yavuz}
\author{Jianguo Zhang}
\affiliation{%
    \country{}
}

%%
%% By default, the full list of authors will be used in the page
%% headers. Often, this list is too long, and will overlap
%% other information printed in the page headers. This command allows
%% the author to define a more concise list
%% of authors' names for this purpose.
\renewcommand{\shortauthors}{Meng, et al.}

%%
%% The abstract is a short summary of the work to be presented in the
%% article.
\begin{abstract}
Dense retrievers have made significant strides in text retrieval and open-domain question answering. However, most of these achievements have relied heavily on extensive human-annotated supervision.
In this study, we aim to develop unsupervised methods for improving dense retrieval models. We propose two approaches that enable annotation-free and scalable training by creating pseudo query-document pairs: query extraction and transferred query generation. 
The query extraction method involves selecting salient spans from the original document to generate pseudo queries. On the other hand, the transferred query generation method utilizes generation models trained for other NLP tasks, such as summarization, to produce pseudo queries.
Through extensive experimentation, we demonstrate that models trained using these augmentation methods can achieve comparable, if not better, performance than multiple strong dense baselines. Moreover, combining these strategies leads to further improvements, resulting in superior performance of unsupervised dense retrieval, unsupervised domain adaptation and supervised fine-tuning, benchmarked on both BEIR and ODQA datasets\footnote{Code and datasets will be publicly available at \url{https://github.com/salesforce/AugTriever}.}.
\end{abstract}

%%
%% The code below is generated by the tool at http://dl.acm.org/ccs.cfm.
%% Please copy and paste the code instead of the example below.
%%


%%
%% Keywords. The author(s) should pick words that accurately describe
%% the work being presented. Separate the keywords with commas.

%% A "teaser" image appears between the author and affiliation
%% information and the body of the document, and typically spans the
%% page.

% \received{20 February 2007}
% \received[revised]{12 March 2009}
% \received[accepted]{5 June 2009}

%%
%% This command processes the author and affiliation and title
%% information and builds the first part of the formatted document.
\maketitle

\section{Introduction}

Text retrieval is currently one of the most influential artificial intelligence applications. Through common internet services like web search and product search, billions of users access vast amounts of data on the Internet, benefiting from information retrieval techniques. While traditional lexical retrieval remains a simple yet effective solution, neural network-based models, particularly dense retrievers, have made significant advancements in recent years, showcasing their advantages in scenarios that involve semantic matching.

However, the majority of dense retrievers heavily depend on training with a large volume of annotated data. For example, MS MARCO~\cite{msmarco} and Natural Questions~\cite{nq} are the two most widely used datasets, and models trained on these datasets have achieved exceptional performance. Nevertheless, each of these datasets comprises hundreds of thousands of query-document pairs annotated by humans, making the collection process prohibitively expensive, and the models trained on them may not generalize well to unseen domains~\cite{beir}. The challenge of training dense retrieval models without human-annotated data remains unsolved.

Recent efforts have shown promising results in training dense retrievers in an annotation-free manner~\cite{contriever,spider,cpt-paper}. Following the conventional paradigm of self-supervised learning, a pretext task is designed by considering two different views of a single document as a positive pair. Subsequently, a dual-encoder model is trained using contrastive learning, aiming to map the two views of the data to similar hidden representations. However, when directly applied on downstream retrieval tasks, these unsupervised models tend to perform worse than the classic method BM25. Nevertheless, a performance boost is observed when fine-tuning the models with annotated positive pairs. This observation motivates us to investigate the gap between the pretext task and downstream retrieval tasks. 

Existing strategies for constructing positive pairs are often heuristic in nature. For example, Contriever~\cite{contriever} randomly samples two text spans from a document to form a positive pair. It is evident that the quality of the positive pairs is poorly controlled, and the resulting pseudo queries bear little resemblance to real-world queries. Consequently, the models are adversely affected by the noisy pseudo pairs, leading to inferior performance on down-stream tasks.

In this study, we propose two novel strategies for constructing pseudo query-document pairs without any retrieval related supervisions. We summarize our contributions as follows:

1. We introduce query extraction (\textsc{\textbf{QExt}}), a novel data augmentation method for training dense retrievers. Given a document, we sample a list of random spans and utilize various techniques to determine their salience. The spans with the highest scores are selected as pseudo queries.

2. We propose transferred query generation (\textsc{\textbf{TQGen}}), where pseudo queries are generated using generation models trained for other NLP tasks, such as summarization, unlike previous studies that heavily rely on human annotated data for training query generation models. To the best of our knowledge, this is the first study demonstrating that the inductive bias from other NLP tasks can be leveraged for training dense retrievers.

3. We contribute two datasets, namely \textsc{\textbf{AugQ-Wiki}} and \textsc{\textbf{AugQ-CC}}, which consist of 22.6M and 52.4M pseudo query-document pairs for unsupervised retrieval training. % These datasets are created by applying our proposed augmentation methods to two large corpora.

4. Extensive experiments show that retrievers using \textsc{\textbf{QExt}} and \textsc{\textbf{TQGen}}, referred to as \textbf{\augtrv}, achieve superior performance and beat strong baselines on BEIR and open-domain QA benchmarks. The results showcase the effectiveness of the proposed augmentation methods as means for retrieval pretraining and domain adaptation, without the need for any human-annotated queries/questions.

\section{Background}
\subsection{Bi-encoder Dense Retriever}
We employ a Transformer based bi-encoder architecture and contrastive learning to train our dense retrievers~\cite{dpr,ance,contriever}. Specifically, we utilize two transformers, denoted as $E_q$ and $E_d$, to encode queries $\mathit{q}$ and documents $\mathit{d}$, respectively. These decoders generate low-dimensional vectors by performing average pooling over the output embeddings of the top layer. The similarity score between $\mathit{q}$ and $\mathit{d}$ is computed by taking the inner product of the two vectors. The encoder parameters are initialized with BERT-base~\cite{bert} and are shared between the encoders. 

The model is optimized using a contrastive objective, where other documents in the same batch are treated as negative examples. Alternatively, recent works~\cite{contriever,xmoco,laprador} use a momentum encoder and a large vector queue to enable the use of additional negative examples. We refer to the architecture that utilizes negative examples in the same batch as \textsc{\textbf{InBatch}}, while the architecture with a momentum document encoder is referred to as \textsc{\textbf{MoCo}}.

\subsection{Construction of Pseudo Query-Document Pairs for Unsupervised Text Retrieval}
\label{sec:unsup-method}
Various methods have been proposed to construct pseudo query-document pairs for training unsupervised dense retrievers. We summarize some of these methods below, while additional related methods can be found in references~\cite{shen2022lowresource,zhao2022dense}.

\begin{itemize}
  \item \textsc{Inverse Cloze Task}~\cite{ict}: A sentence is randomly selected from a given document, and a retriever is trained to retrieve the document using the sentence as a query.
  \item Masked salient span in \textsc{REALM}~\cite{realm}: REALM is a retrieval-augmented language model. During its pre-training phase, a retriever and a generator work together to predict a masked named entity. 
  \item Random cropping (\textsc{RandomCrop}) in \contrv~\cite{contriever}: For a given document \textit{d}, two random spans (contiguous subsequences) are independently extracted from \textit{d} to create a positive pair.
  \item \spider~\cite{spider}: This method selects two passages within a document that contain identical n-grams (recurring spans) as a positive pair. It should be noted that this method may not be as data-efficient, as recurring spans may not be present in all documents, particularly shorter ones.
  \item \cpt~\cite{cpt-paper}: Positive pairs are constructed by using neighboring text pieces from the same document.
  \item \spar~\cite{spar}: The dense lexical model $\Lambda$ is trained with questions or random sentences as queries, paired with the top K passages retrieved by BM25.
  
  % It is trained to mimic the ranking of a sparse teacher retriever BM25 using contrastive learning.
\end{itemize}

\begin{figure*}[!h]
\centering
\includegraphics[width=0.8\textwidth]{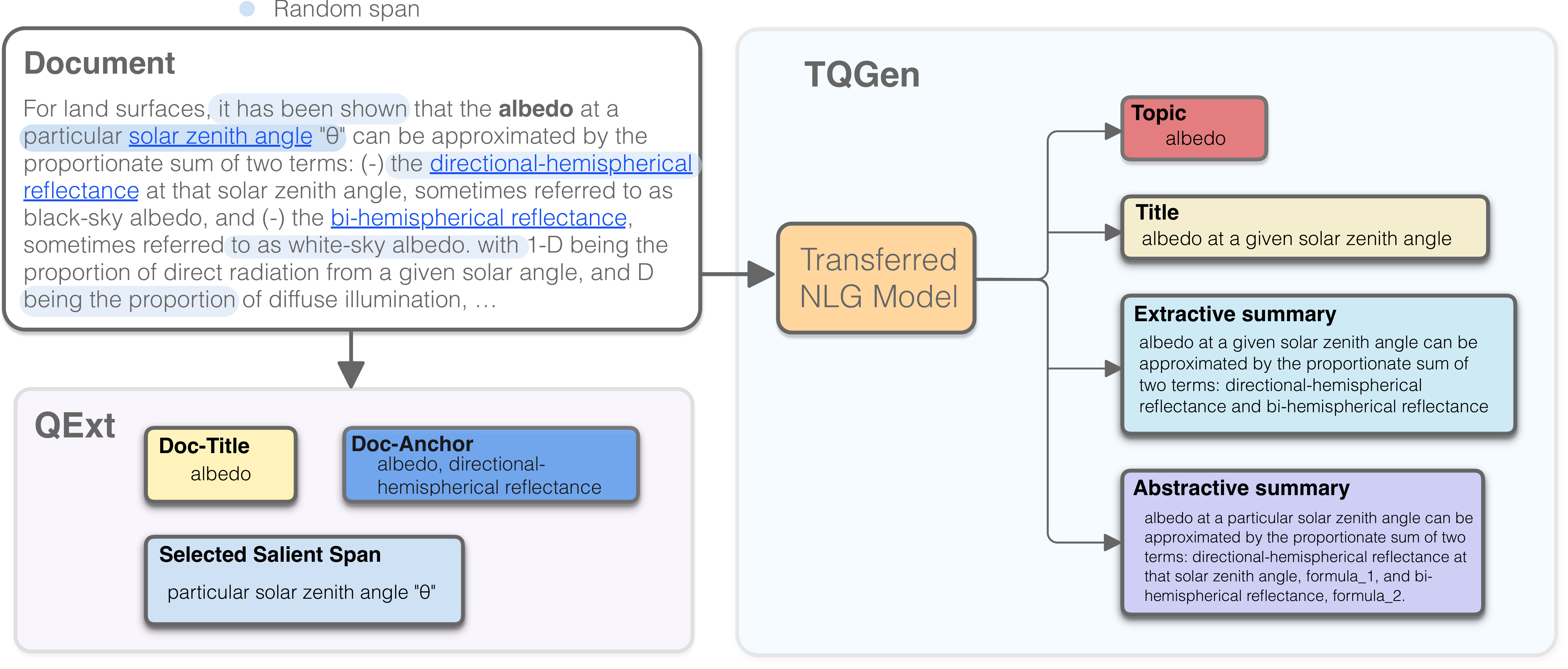}
\caption{An overview of proposed augmentation methods for \augtrv.}
\label{normal_case}
\end{figure*}

\section{Method}
\label{sec:method}
In this section, we present several data argumentation methods for generating pseudo queries from a given document without the need for annotated queries or questions. We apply those methods on Wikipedia passages and CommonCrawl web documents, resulting in two large augmented datasets called \textsc{AugQ}. Subsequently, we train bi-encoder dense retrievers using \textsc{AugQ} and refer to the resulting models \augtrv. These models are trained using either \textsc{InBatch} or \textsc{MoCo}.

\subsection{Query Extraction (\textsc{QExt})}
Given a document, we hypothesize that certain parts of it contain more representative information. Therefore, we extract and utilize these parts as pseudo queries to train the retrievers. 

\subsubsection{Query Extraction Using Document Structural Heuristics}
Documents often have rich structures, and extracting information based on these structures for weak supervision has been shown to be effective~\cite{chang2019ICT+WLP+BFS,zhou2022hyperlink,wu2022hyperlink}.
Following this line of research, we propose utilizing the document structure to construct weakly annotated queries for training the dense retriever. Specifically, we consider two types of information, \textit{titles} (\textsc{Doc-Title}) and \textit{anchor texts} (\textsc{Doc-Anchor}). Titles and anchor texts are similar to search queries and are commonly available on the internet. They are considered representative of the core content of the document by the document authors. Extracting titles and anchor texts can be achieved using DOM structures and human-crafted heuristics.

\subsubsection{Query Extraction Using Salient Span Selection}
The previous method heavily relies on the quality and availability of distant labels embedded in the document structures, which may limit the scalability of training. To address this limitation, we propose an alternative approach that directly extracts informative spans from a document. The hypothesis is that a document can be segmented into multiple spans, and some of these spans are more representative than others. We can then employ various methods to select the most salient spans as pseudo queries. It is important to note that we do not mask the selected spans in the document.

Formally, given a document $\textit{d}$, we randomly sample a number of text spans $s_{1}, s_{2}, ..., s_{N}$ from it. We consider 16 random spans, with lengths ranging from 4 to 16 words. We propose three approaches to measure the salience between $\textit{d}$ and each of these spans:

\begin{itemize}
  \setlength\itemsep{0.1em}
  \item \textsc{QExt-Self}: This method selects spans by leveraging the model itself. We input each span $s_{i}$ paired with $\textit{d}$ into the model and use the dot-product as the salience score.
  \item \textsc{QExt-BM25}: This method selects spans based on lexical models. BM25 is a widely used method for measuring the lexical relevance between queries and documents. Here, we utilize BM25 to select spans based on their lexical statistics.
  \item \textsc{QExt-PLM}: This method selects spans using pre-trained language models (PLMs). PLMs have shown remarkable performance in text understanding and generation tasks. In our approach, we utilize PLMs to measure the relevance by assessing how likely a span can be generated given a document as the context. Specifically, we feed the document as a prefix to a T5-small LM-Adapted model~\cite{t5} and use the likelihood $p(s_{i}|\textit{d})$ as the salience score for the span.
\end{itemize}

\subsection{Transferred Query Generation (\textsc{TQGen})}
Previous studies have demonstrated the effectiveness of query generation as a means of augmenting training data~\cite{paq,doc2query}. However, these approaches typically require a significant amount of annotated data to train a query generator. In our work, we propose a different approach, utilizing text generation models of irrelavant tasks to produce pseudo queries as distant supervision. We hypothesize that the inductive bias of these tasks can effectively bootstrap the training of dense retrievers. Specifically, we leverage models for summarization and keyphrase generation, as these outputs are commonly considered relevant and representative of the source text. Other options include paraphrasing or back-translation.

For implementation, we use a single \textsc{T0} model (3B parameters) as a meta generator, eliminating the need for selecting models for each generation task. We provide \textsc{T0} with various prompts to generate outputs for different tasks, including:
\begin{itemize}
  \setlength\itemsep{-0.0em}
  \item \textsc{TQGen-Topic}: \textit{What is the main topic of the text above?}
  \item \textsc{TQGen-Title}: \textit{Please write a title of the text above}.
  \item \textsc{TQGen-AbSum} (Abstractive summary): \textit{Please write a short summary of the text above}.
  \item \textsc{TQGen-ExSum} (Extractive summary): \textit{Please use a sentence from the above text to summarize its content}.
\end{itemize}
We intentionally include a prompt for extractive summaries to encourage the model to use words from the original text and reduce the risk of hallucination. We employ nucleus sampling to generate a single pseudo query for each document, with parameters Top-p=0.9 and Top-K=0. We explore two hybrid settings by combining multiple strategies: \textit{Hybrid-All} which uses all proposed strategies, and \textit{Hybrid-TQGen} which mainly uses \textsc{TQGen}.

\section{Experiments}
\subsection{Datasets}
\noindent\textbf{Training Data}:
To generate augmented query-document data (\textsc{AugQ}), we utilize two large text datasets: Wikipedia\footnote{enwiki-20211020-pages-articles-multistream.xml.bz2} and CommonCrawl by Pile~\cite{pile} (Pile-CC). For Wikipedia, we process the original text dump by segmenting articles into paragraphs by line breaks and reserving titles and anchor texts (texts with hyperlinks, italics, or boldface). This results in a total of 22.6 million paragraphs available for training. Pile-CC consists of 52.4 million web documents, but it does not provide structure information, making \textsc{Doc-Anchor} unavailable. For \textsc{Doc-Title}, we extract the first line of each document as its title, truncating it to a maximum of 64 words. We manually inspected a few hundred examples and found that it correctly extracted titles in approximately 50\% of the cases.

\noindent\textbf{Test Data}:
We use two benchmarks for evaluation: BEIR~\cite{beir} and six Open-Domain Question Answering (ODQA) datasets. 
We consider BEIR to be a better benchmark for information retrieval as it covers a broader range of domains and a wide variety of query types. We discuss the scores of MS MARCO (MM) separately since it is one of the most extensively studied IR test sets.
On the other hand, all ODQA datasets are based on Wikipedia and primarily designed for evaluating question answering systems. Therefore, they may introduce certain domain and task biases. We utilize these datasets for retrieval evaluation following previous studies~\cite{dpr,spider}. We report scores on SQuAD v1.1 (SQ) and EntityQuestions (EQ) separately, as they tend to favor lexical models, while the other four datasets may favor semantic matching approaches.

% \noindent\textbf{BEIR}:
% We use BEIR~\cite{beir} benchmark to test models' retrieval performance across domains. BEIR is a collection of 18 testing datasets from 9 heterogeneous retrieval tasks. Following previous studies, we mainly use 14 public datasets plus MS MARCO~\cite{msmarco}. Note that differing from most studies that fine-tune models on MS MARCO and zero-shot evaluate them on other BEIR datasets, all results except for Sec~\ref{sec:finetune} do not use any annotated data.
% \noindent\textbf{Open-Domain Question Answering (ODQA)}:
%  We also test all models for retrieval only on six Wikipedia-based ODQA datasets following previous setups~\cite{dpr,spider}: Natural Questions (NQ)~\cite{nq}, TriviaQA (TQA)~\cite{triviaqa}, WebQuestions (WebQ)~\cite{webq}, CuratedTREC (TREC)~\cite{trec}, SQuAD v1.1 (SQ)~\cite{squad} and EntityQuestions (EQ)~\cite{eq}. We use the Wikipedia passages provided by DPR~\cite{dpr}. Note that a different version of NQ is also included in BEIR.

\subsection{Implementation Details}
Our models use either \textsc{\textbf{InBatch}} or \textsc{\textbf{MoCo}} architecture as the backbone, initialized with BERT-base~\cite{bert}.
We adopt most of the training settings used by \contrv~\cite{contriever}.
% but on a significantly smaller scale. For instance, our training utilized 52 million documents from Pile-CC, compared to the 700 million in CCNet used by Contriever. Additionally, our model underwent up to 200,000 training steps, significantly fewer than the 500,000 steps employed by Contriever.
Note that our training setting is considerably smaller (200k steps, trained with 55M documents) compared with \contrv (500k steps, trained with 700M documents from CCNet~\cite{ccnet}). All experiments are conducted on cloud instances equipped with up to 16 NVIDIA A100 GPUs (40GB) and most training jobs were finished within 48 hours.
\subsubsection{Pretraining}
\augtrv is trained with the following configurations:
\begin{enumerate}
\setlength\itemsep{0.1em}
\item Most models are trained for 100k steps.
\item The batch size is set to 1,024 for most models, and 2,048 for \textsc{AugQ-CC} with \textsc{TQGen}/\textsc{QGen}/\textsc{Hybrid}.
\item We use a learning rate (\textit{lr}) of $5e^{-5}$ with 10k steps of linear warmup.
\item The optimizer used is ADAM.
\item For the MoCo architecture, we set the queue size to $2^{14}$ (\contrv used a queue size of $2^{17}$). We empirically observed that a larger queue size can deteriorate the performance.
\item For the hybrid settings: (1) \textit{Hybrid-All}: We use \textsc{RandomCrop} for 20\% of the spans, \textsc{QExt-PLM} for 10\%, and a combination of \textsc{Doc-Title} and \textsc{TQGen} for the remaining 70\%.
(2) \textit{Hybrid-TQGen}: We use \textsc{RandomCrop} for 20\% of the spans and rely solely on \textsc{TQGen} for the remaining 80\%.
\item We experiment with two larger scale training: (1) \textit{Hybrid-TQGen+} is trained for 200k steps using a batch size of 2048 and queue size of $2^{14}$; (1) \textit{Hybrid-TQGen++} is trained for 200k steps using a batch size of 4096 and queue size of $2^{16}$.

\end{enumerate}

\subsubsection{Domain Adapataion}
For domain adaptation, we train a pre-trained model (Hybrid-TQGen++) for up to 2k steps with a batch size of 256/32 (MSMARCO\&Wikipedia/others) and a learning rate of $1e^{-5}$. We train single models for five Wikipedia datasets (NQ, HotpotQA, DBPedia-Entity, FEVER and Climate-FEVER) and two community QA datasets (Quora and CQADupStack), resulting in 10 different models for target domains.
\subsubsection{Fine-tuning}
For fine-tuning, we train models using MSMARCO\footnote{\url{https://huggingface.co/datasets/sentence-transformers/embedding-training-data/blob/main/msmarco-triplets.jsonl.gz}} for 10k steps with a batch size of 1,024, pairing each example with one hard negative example (1 positive + 2,047 negative), using a learning rate of $1e^{-5}$).
For a fair comparison, we fine-tune \spar using the query-encoder only. For all reproduced fine-tuning, we use pooling and vector normalization in consistence with the way in their pretraining.

\subsection{Baselines}
We consider several unsupervised dense methods discussed in Section \ref{sec:unsup-method} as baselines. These include \bm as a lexical baseline, and five dense baselines: \contrv, \spider, \spar (Wikipedia version), LaPraDor (no BM25)~\cite{laprador} and CPT~\cite{cpt-paper}. We report their scores if publically available (BEIR results of \bm and \contrv), or reproduce the results using public code and checkpoints (indicated with $\dagger$). \textsc{MoCo+RandomCrop} can be regarded as our reproduced \contrv in a smaller scale. We also include baselines with generated queries (using a supervised Doc2Query\footnote{\url{https://huggingface.co/doc2query/all-with_prefix-t5-base-v1}}~\cite{doc2query}) and questions PAQ~\cite{paq}, referred to as \textsc{QGen-D2Q} and \textsc{QGen-PAQ} respectively. We rerun most baselines on Touché-2020 (v2) since the data has been updated in BEIR.

\subsection{Results}

\begin{table}[!tb]
\small
\caption{Unsupervised retrieval performance (MM/BEIR nDCG@10 and ODQA Recall@20). MM denotes scores on MS MARCO. QA4 denotes averaged scores of NQ, TQA, WebQ and TREC. We highlight the \colorbox{BrickRed!30}{best} and \colorbox{RoyalBlue!20}{second best} in each column, and \textbf{best} in each group per column.}

\begin{adjustbox}{width=0.48\textwidth}
\renewcommand{\arraystretch}{0.85}
\begin{tabular}{L{2.4cm}lcccccc}
\toprule
\rowcolor{white}
\textbf{Group} & \textbf{Model} & \textbf{MM} & \textbf{BEIR14} & \textbf{CPT-subset} & \textbf{QA4} & \textbf{SQ\&EQ} \\
\midrule
\multirow{7}{*}{Baseline}    & BM25       & \textbf{22.8} & \bestcell\textbf{43.0} & \secbestcell\textbf{46.1} & 70.7$\dagger$ & \bestcell\textbf{71.3}$\dagger$ \\
                             & LaPraDor            & 16.9 & 30.2 & 33.6 & - & - \\
                             & SPAR$\dagger$       & 19.3 & 37.3 & 41.4 & 69.1 & 67.7 \\
                             & Spider$\dagger$     & 15.0 & 28.1 & 31.3 & 73.0 & 63.6 \\
                             & Contriever          & 20.6 & 36.9 & 40.7 & \textbf{73.1}$\dagger$ & 63.9$\dagger$ \\
                             & CPT-text S          &-&-& 42.2 &-&- \\
                             & CPT-text M          &-&-& 43.2 &-&- \\
                             & CPT-text L          &-&-& 44.2 &-&- \\
% \cdashline{1-4}[2.5pt/5pt]

\midrule
\rowcolor{Gray}
                                   & InBatch    & 14.0 & 25.3 & 28.0 & 61.9 & 44.9 \\
\rowcolor{Gray}
\multirow{-2}{*}{RandomCrop(Wiki)}  & MoCo      & 17.5 & 30.9 & 34.1 & 64.6 & 52.5 \\

\rowcolor{white}
                                    & InBatch   & 16.3 & 27.4 & 30.6 & 73.2 & 57.7 \\
\rowcolor{white}
\multirow{-2}{*}{RandomCrop(CC)}     & MoCo     & 19.2 & 34.0 & 37.5 & 71.5 & 61.9 \\

\rowcolor{Gray}
                                   & InBatch    & \bestcell\textbf{25.4} & 38.5 & 42.5 & 78.6 & 67.1 \\
\rowcolor{Gray}
\multirow{-2}{*}{QGen-D2Q(Wiki)}   & MoCo       & 23.7 & 38.5 & 42.3 & 77.4 & 67.4 \\

\rowcolor{white}
                                   & InBatch    & 24.4 & 39.5 & 42.8 & 75.6 & 63.2 \\
\rowcolor{white}
\multirow{-2}{*}{QGen-D2Q(CC)}     & MoCo       & 23.2 & \textbf{39.8} & \textbf{43.7} & 76.6 & 65.6 \\

\rowcolor{Gray}                    & InBatch    & 22.3 & 31.0 & 32.9 & \bestcell\textbf{81.3} & \secbestcell \textbf{70.9} \\
\rowcolor{Gray}
\multirow{-2}{*}{QGen-PAQ (Wiki)}   & MoCo       & 22.6 & 29.9 & 32.6 & \secbestcell 78.7 & 68.9 \\

\midrule
\multicolumn{4}{l}{\textbf{\augtrv-CC}}\\
\midrule
\rowcolor{Gray}
                               & InBatch    & 19.7 & 33.2 & 36.5 & 73.5 & 59.7 \\
\rowcolor{Gray}
\multirow{-2}{*}{Doc-Title}    & MoCo       & 21.8 & 38.7 & 42.7 & 74.8 & 64.3 \\

\multirow{2}{*}{QExt-BM25}   & InBatch    & 16.3 & 27.8 & 30.7 & 73.5 & 58.3 \\
                             & MoCo       & 20.2 & 35.8 & 39.6 & 73.1 & 63.7 \\
\rowcolor{Gray}
                               & InBatch    & 16.2 & 27.2 & 40.4 & 73.4 & 57.9 \\
\rowcolor{Gray}
\multirow{-2}{*}{QExt-PLM}      & MoCo       & 20.6 & 38.2 & 42.3 & 73.0 & 64.1 \\

                               & InBatch    & 20.7 & 39.0 & 43.0 & 71.6 & 60.5 \\
\multirow{-2}{*}{TQGen-Topic}  & MoCo       & 21.2 & 38.9 & 43.1 & 73.3 & 63.4 \\

\rowcolor{Gray}
                               & InBatch    & 20.0 & 37.8 & 41.3 & 72.5 & 60.0 \\
\rowcolor{Gray}
\multirow{-2}{*}{TQGen-Title}   & MoCo       & 21.8 & 39.3 & 43.3 & 74.2 & 64.2 \\

\rowcolor{white}
                              & InBatch    & 18.1 & 35.3 & 38.6 & 72.0 & 57.4 \\
\rowcolor{white}
\multirow{-2}{*}{TQGen-AbSum}  & MoCo       & \textbf{23.2} & \textbf{39.6} & \textbf{43.5} & 74.4 & \textbf{64.9} \\
\rowcolor{Gray}
                               & InBatch    & 18.9 & 36.3 & 39.6 & 72.8 & 58.7 \\
\rowcolor{Gray}
\multirow{-2}{*}{TQGen-ExSum}  & MoCo       & 23.0 & 39.4  & 43.4 & \textbf{74.8} & \textbf{64.9} \\

\midrule
\multicolumn{4}{l}{\textbf{\augtrv-CC with Hybrid Strategies}}\\
\midrule
% Hybrid-TQGen  & MoCo    & 23.3 & 37.8 & 41.5 & 75.5 & 64.8 \\
\rowcolor{Gray}
Hybrid-All    & MoCo    & 23.5 & 39.4 & 43.3 & 74.1 & 64.4 \\
Hybrid-TQGen  & MoCo    & 23.3 & 39.0 & 43.7 & 74.3 & 64.4 \\
\rowcolor{Gray}
Hybrid-TQGen+  & MoCo    & \secbestcell\textbf{24.6} & 41.1 & 45.2 & 76.0 & 65.9 \\
Hybrid-TQGen++ & MoCo    & \bestcell\textbf{25.4} & \secbestcell\textbf{42.1} & \bestcell\textbf{46.2} & \textbf{76.2} & \textbf{67.1} \\

\midrule
\multicolumn{4}{l}{\textbf{\augtrv-Wikipedia}}\\
\midrule
\rowcolor{Gray}
                             & InBatch    & 15.6 & 29.8 & 33.2 & 64.8 & 49.9 \\
\rowcolor{Gray}
\multirow{-2}{*}{Doc-Anchor} & MoCo       & 17.9 & 35.4 & 39.2 & 68.5 & 57.4 \\

\rowcolor{white}
                             & InBatch    & 14.7 & 30.0 & 33.9 & 61.9 & 52.1 \\
\rowcolor{white}
\multirow{-2}{*}{Doc-Title}  & MoCo       & 18.5 & 33.7 & 37.1 & 68.6 & 58.4 \\

\rowcolor{Gray}
                             & InBatch    & 15.0 & 26.3 & 28.2 & 61.5 & 43.8 \\
\rowcolor{Gray}
\multirow{-2}{*}{QExt-PLM}    & MoCo       & 18.6 & 34.3 & 37.8 & 66.6 & 55.7 \\
\rowcolor{white}
                             & InBatch    & 21.3 & \textbf{38.9}  & \textbf{43.2} & 72.4 & 64.4 \\
\rowcolor{white}
\multirow{-2}{*}{TQGen-Topic}  & MoCo       & 21.3 & 38.3 & 42.5 & 73.6 & 65.3 \\
% \multirow{2}{*}{TQGen-Title} & InBatch    & 20.8 & 38.8 & 42.9 & 74.0 & 64.6 \\
%                              & MoCo       & \textbf{21.6} & 38.5 & 42.8 & \textbf{74.1} & 64.9 \\

\rowcolor{Gray}
                              & InBatch    & 17.4 & 36.3 & 40.2 & 74.9 & 65.3 \\
\rowcolor{Gray}
\multirow{-2}{*}{TQGen-AbSum}  & MoCo       & 21.2 & 37.2 & 41.3 & 74.5 & 65.4 \\

\rowcolor{white}
                             & InBatch    & 18.2 & 36.7 & 40.2 & \textbf{75.6} & 65.4 \\
\rowcolor{white}
\multirow{-2}{*}{TQGen-ExSum} & MoCo       & \textbf{22.5} & 37.9 & 41.8 & 75.1 & \textbf{66.7} \\

%\midrule
% Doc-Title(Pile6)             & MoCo       & 22.2 & \textbf{39.6} & & \textbf{74.3} & 62.8 \\
% Doc-Title(Pile10)            & MoCo       & 21.4 & 39.4 & & 73.8 & 63.1 \\
\bottomrule
\end{tabular}
\end{adjustbox}
\label{tab:main-result}
\end{table}
\vspace{-4px}

\subsubsection{Unsupervised Retrieval}
\label{sec:main-result}
% A previous study~\cite{contriever} has shown that dense retrievers can benefit from training with large and domain-general data. We also explore this hypothesis by scaling the training with Pile-CC, a large corpus of web documents. We adopt a larger batch size of 2,048 for \textsc{TQGen} and \textsc{QGen} with MoCo. We separate the scores of SQuAD (SQ) and EntityQuestions (EQ) from the rest four ODQA datasets, as they favor lexical models like \bm.
We present the main unsupervised results in Table~\ref{tab:main-result} and will discuss certain details in Sec~\ref{sec:analysis}. Among all unsupervised baselines, \bm still outperforms the other baselines by a significant margin. For dense retrievers, the lexical-oriented retriever \spar performs the best on BEIR14 and SQ\&EQ, indicating that dense retrievers can achieve robust retrieval performance through a lexical teacher. \contrv performs comparably with \spar on BEIR.
% , and it outperforms the others on MM and QA4.
% CPT-text outperforms other dense models on BEIR with the advantage of scale. 
The supervised augmentation \textsc{QGen-D2Q} delivers competitive results on both benchmarks, suggesting that query generation trained with MS MARCO can work well both in-domain and out-of-domain.
% Particularly, the one trained on Wikipedia excels on MS MARCO and ODQA, and the CC variant demonstrates great generalization ability on BEIR.

Regarding \augtrv\footnote{We train InBatch models using augmented query-document pairs only, whereas we train MoCo models using a 50/50 mixed strategy (50\% of pairs by \textsc{RandomCrop} and 50\% by one of the proposed augmentation strategies). Further comparisons between the two settings are discussed in Sec~\ref{sec:analysis}.}, we find that multiple variants, e.g. MoCo+QExt-PLM and TQGen models, significantly outperform dense baselines on BEIR, especially when trained with the domain-general data CC. 
% Additionally, multiple variants achieve better scores than \bm on MM and QA4. 
Notably, our best results are achieved by hybrid strategies and longer training (Hybrid-TQGen+/++). Hybrid-TQGen++ outperforms CPT-text L on BEIR by a large margin, despite CPT-text L being 20 times larger than \augtrv models. These empirical results strongly suggest the effectiveness of the proposed method for unsupervised dense retrieval.

Besides, (1) we observe that \textsc{TQGen} achieves the overall best performance, indicating that the outputs of transferred NLP tasks, such as keyword and summary generation~\cite{meng2017deep,see2017get}, can be utilized for training dense retrieval models effectively; (2) It is worth noting that \textsc{TQGen-topic} generalizes well under all settings, suggesting that keywords can serve as robust surrogate queries. (3) \textsc{MoCo+QExt-PLM} outperforms all dense baselines on BEIR, indicating that query extraction can be an effective unsupervised method. However, since it scores random spans using an LM on-the-fly, we are unable to scale it up (larger batch size, better scorer) in this study.

\subsubsection{Unsupervised Domain-Adaptation}
\label{sec:domain-adapt}

\begin{table*}[ht]
\fontsize{7}{6.5}\selectfont
\renewcommand{\arraystretch}{1.1}

\setlength{\tabcolsep}{2pt}

\caption{Results of domain-adapation on 14 BEIR testsets. We highlight the \colorbox{BrickRed!30}{best} and \colorbox{RoyalBlue!20}{second best} scores.}
\vspace{-1em}
\begin{adjustbox}{width=1.0\textwidth}
\begin{tabular}{lcccccccccccccccc}
\toprule
& \textbf{BEIR14} & \textbf{TREC-COVID} & \textbf{NFCorpus} & \textbf{NQ} & \textbf{HotpotQA} & \textbf{FiQA} & \textbf{ArguAna} & \textbf{Touche} & \textbf{DBPedia} & \textbf{Scidocs} & \textbf{FEVER} & \textbf{Cli-FEVER} & \textbf{Scifact} & \textbf{Quora} & \textbf{CQADup} \\

\midrule
\multicolumn{2}{l}{\textbf{Unsupervised}} \\
BM25                                      & 43.0 & 65.6 & 32.5 & 32.9 & 60.3 & 23.6 & 31.5 & \bestcell36.7 & 31.3 & 15.8 & \secbestcell 75.3 & 21.3 & 66.5 & 78.9  & 29.9 \\
COCO-DR$_{Base}$ & 29.5 & 43.9 & 23.7 & 9.9  & 23.4 & 17.5 & 43.1 & 11.7 & 15.6 & 11.2 & 25.9 & 8.9  & 71.1 & 79.3 & 27.6 \\
TSDAE                                     & 43.6 & 70.8 & 31.2 & \bestcell47.1 & \bestcell63.8 & 29.3 & 37.5 & 21.8 & 35.4 & 15.4 & 64.0 & 16.8 & 62.8 & 83.3 & 31.8 \\
\midrule
\multicolumn{2}{l}{\textbf{MS MARCO involved}} \\
TAS-B                                     & 42.7 & 48.5 & 31.9 & 46.3 & 58.4 & 29.8 & 43.4 & 16.2 & \bestcell38.4 & 14.9 & 69.5 & 22.1 & 63.5 & 83.5 & 31.5 \\
QGen                                      & 42.1 & 56.0 & 31.4 & 35.4 & 51.4 & 28.7 & \secbestcell 52.4 & 17.1 & 33.1 & 15.5 & 63.8 & 22.5 & 63.8 & 85.0 & 33.0 \\
TSDAE+QGen                                & 43.4 & 58.4 & 33.7 & 34.6 & 52.2 & 31.4 & \bestcell 54.7 & 17.2 & 33.2 & \secbestcell 17.1 & 64.2 & 22.6 & 66.7 & \bestcell85.7 & \bestcell35.3 \\
TAS-B+QGen                                & 42.8 & 56.6 & 33.4 & 36.3 & 52.0 & 30.1 & 51.8 & 17.5 & 32.7 & 16.4 & 63.9 & \bestcell 24.4 & 65.3 & 85.3 & 33.7 \\
DocT5-Query                               & 45.3 & \secbestcell 71.3 & 32.8 & 39.9 & 58.0 & 29.1 & 46.9 & \secbestcell 34.7 & 33.1 & 16.2 & 71.4 & 20.1 & 67.5 & 80.2 & 32.5 \\
GPL                                       & \bestcell 46.5 & \bestcell 71.8 & 34.2 & \secbestcell 46.7 & 56.5 & \secbestcell 32.8 & 48.3 & 23.1 & \secbestcell 36.1 & 16.1 & \bestcell 77.9 & \secbestcell 22.7 & 66.4 & 83.2 & \secbestcell 34.5 \\

% TQGen-Topic & 39.6   & 61.3       & 32.1     & 27.9  & 51.7     & 25.6   & \textbf{39.4}    & 21.0        & 30.3    & 15.1    & 62.5   & 15.2          & \textbf{66.8}    & 78.1  & 27.6        \\
% \multicolumn{1}{r}{\textsc{+DA}}  & 41.1   & \textbf{71.0}       & 31.1     & 28.2  & 53.1     & \textbf{33.6}   & 18.4    & 19.4        & \textbf{32.3}    & \textbf{17.3}    & 72.9   & 16.1          & 64.8    & \textbf{80.8}  & \textbf{35.8}        \\
% \multicolumn{1}{r}{\textsc{diff\%}}  & 3.6\%  & 15.7\%     & -3.1\%   & 1.3\% & 2.7\%    & 31.4\% & -53.2\% & -7.8\%      & 6.7\%   & 14.4\%  & 16.7\% & 5.6\%         & -3.0\%  & 3.5\% & 29.7\%      \\
\midrule
\textbf{Ours (Unsupervised)} \\
Hybrid-TQGen++                           & 42.1 & 64.0  & \secbestcell 35.1  & 33.5  & 57.6  & 29.2   & 39.8   & 19.8   & 32.7  & 16.7  & 65.3   & 15.9  & \secbestcell 68.7 & 79.9  & 30.6   \\
\multicolumn{1}{r}{\textsc{+DA}}         & \secbestcell 45.4 & 68.1  & \bestcell36.8  & 33.7  & \secbestcell 60.8  & \bestcell34.0   & 46.2   & 23.3   & 34.9  & \bestcell17.2  & 72.0   & 18.4  & \bestcell70.3  & \secbestcell 85.4  & \secbestcell 34.5   \\
\multicolumn{1}{r}{\textsc{diff\%}}      & 8.0\% & 6.4\% & 4.8\% & 0.7\% & 5.6\% & 16.3\% & 16.3\% & 17.6\% & 6.9\% & 2.4\% & 10.3\% & 15.5\% & 2.4\% & 6.9\% & 12.7\% \\

\bottomrule
\end{tabular}
\end{adjustbox}

\label{tab:domain-adapt}
\vspace{-1em}
\end{table*}

Retrieval models are often applied to data of new domains, making domain adaptation crucial in real-world scenarios. We investigate the effectiveness of the proposed augmentation methods for domain adaptation. 
% We use an InBatch model pre-trained with \textsc{AugQ-CC} and \textsc{TQGen-Topic} and fine-tune it with in-domain pseudo query-document pairs. 
We leverage \textsc{TQGen-Topic} method for its simplicity and overall great performance in Table~\ref{tab:main-result}. We generate main topics (keywords) for documents in each BEIR domain (test set), and then fine-tune Hybrid-TQGen++ with the in-domain pseudo query-document pairs. We compare our model with various baselines reported by \citeauthor{beir,gpl,yu2022coco}. Note that several baselines use MS MARCO pairs for training models (TAS-B), query generators (QGen and GPL) and rerankers (GPL), while in contrast our models have not used any retrieval related data. 

The results are presented in Table~\ref{tab:domain-adapt}. We observe an 8\% average gain over 14 BEIR datasets, suggesting the importance of adapting models using in-domain documents. Significant improvements (up to 16\%) are seen on domains that are specific and distant from the pretraining distribution, such as finance (FiQA) and argument (Tóuche-2020, ArguAna). Furthermore, our model, in spite of the simple method being used, outperforms BM25 and most neural domain adaptation methods. GPL outruns our method with the advantage of using MS MARCO trained query generator and reranker, whereas our models merely use pseudo keywords for training, presenting a simple yet effective approach for domain adaptation. Here we did not explore how augmentation strategies other than \textsc{TQGen-Topic} would work on DA, and we think the pair filtering used by GPL can also help our models. We leave them for future work.

% Out of 15 BEIR datasets, domain adaptation leads to positive impacts on 11 of them, and it outperforms BM25 on 7 datasets (only 3 if w/o DA). The largest improvements are observed in domains that are specific and distant from the pretraining distribution, such as finance (FiQA) and science (COVID, Scidocs). However, negative impacts are observed in four domains where only a small number of documents are available. With the exception of Touché-2020, the remaining three test sets have no more than 10k documents, which may have resulted in overfitting during the fine-tuning process. Additional in-domain data, regularization techniques, and hyperparameter tuning could be beneficial, and we leave them for future work.

\subsubsection{Fine-Tuning with MS MARCO}
\label{sec:finetune}

To assess the effectiveness of the proposed augmentation methods as pretraining measures, we present the fine-tuned results on MS MARCO in Table~\ref{tab:finetune}. We use basic fine-tuning settings without employing advanced techniques such as negative mining~\cite{contriever} or asynchronous index refresh~\cite{ance}. We compare with multiple baselines reported in BEIR~\cite{beir} (BM25, DPR~\cite{dpr}, ANCE, ColBERT~\cite{colbert} ) and we fine-tune the other pretrained models under the same setting (Spider, LaPraDor, Condenser~\cite{condenser}, CoCondenser~\cite{cocondenser}, and Contriever).

Among all baselines, ColBERT and Contriever perform best on BEIR overall, indicating the benefit of late-interaction and extensive pre-training. Most \augtrv models demonstrate equal or better performance to baselines, indicating the effectiveness of using \textsc{AugQ} for pretraining. Our best model, Hybrid-TQGen++, achieves best score on BEIR (45.8), showing strong generalization performance when zero-shot transferred on 14 different datasets. It falls behind \contrv on MS MARCO by a small margin (0.4 point), and we think it can be attributed to the fact that \contrv was pre-trained with both Wikipedia and CommonCrawl data (Hybrid-TQGen++ scores 50.3 on five Wikipedia related datasets, comparing to \contrv's score 51.1), and its training duration was longer (ours 200k steps vs. \contrv 500k steps).
In most cases, the trends in the fine-tuned scores align with the unsupervised results, providing strong evidence that the inductive bias from various augmentation methods can benefit downstream retrieval tasks. 
% The InBatch model outperforms its MoCo counterpart on MS MARCO and BEIR, but lags behind on ODQA datasets. Among all the variants, \textsc{InBatch+TQGen-Topic} achieves the highest performance, followed by other \textsc{TQGen} variants, \textsc{Doc-Title}, and \textsc{QExt-PLM}. The relative advantage of MoCo observed in the unsupervised setting is not consistently carried over to the fine-tuned results, possibly due to the architectural differences between pretraining and fine-tuning stages.

\begin{table}[!h]
% \fontsize{7}{6.5}\selectfont
\tiny
\renewcommand{\arraystretch}{0.6}

\caption{Retrieval scores after fine-tuning with MS MARCO. \augtrv models are pre-trained on \textsc{AugQ-CC} using MoCo. We highlight the \colorbox{BrickRed!30}{best} and \colorbox{RoyalBlue!20}{second best} in each column. $\dagger$ indicates results by us, fine-tuned using public checkpoints.}
\vspace{-1em}
\begin{adjustbox}{width=0.35\textwidth}
\begin{tabular}{lcc}
\toprule
\multicolumn{1}{l}{\textbf{Model}} & \textbf{MM} & \textbf{BEIR14}\\
\midrule
\textbf{Baseline}\\
\midrule
BM25                                     & 22.8 & 43.0                 \\
DPR                                      & 35.4 & 36.8                 \\
ANCE                                     & 38.8 & 40.5                 \\
% GTR-base                        & 42.0 & 44.1                 \\
ColBERT                                  & 40.1 & 44.4                 \\
Spider$\dagger$                          & 24.8 & 19.1                 \\
LaPraDor$\dagger$                        & 38.9 & 40.4                 \\
SPAR$\dagger$                            & 38.0 & 41.5                 \\
Condenser(Book\&Wiki)$\dagger$           & 38.7 & 40.9                 \\
CoCondenser(MSMARCO)$\dagger$            & 40.8 & 42.9                 \\
Contriever$\dagger$                      & \bestcell \textbf{41.3} & \secbestcell \textbf{45.2}                 \\
\midrule
\textbf{Ours}\\
\midrule
QGen-D2Q                        & 39.6 & 43.9                 \\
RandomCrop                      & 38.4 & 42.4                 \\
\hdashline
QExt-PLM                        & 38.8 & 42.5                 \\
TQGen-Topic                     & 38.8 & 43.2                 \\
TQGen-Title                     & 38.8 & 43.2                 \\
TQGen-AbSum                     & 39.1 & 43.9                 \\
TQGen-ExSum                     & 38.8 & 43.1                 \\
\hdashline
Hybrid-All                     & 38.9 & 43.6                 \\
Hybrid-TQGen+                  & 40.3 & 44.7                 \\
Hybrid-TQGen++                 & \secbestcell \textbf{40.9} & \bestcell \textbf{45.8}                 \\

\bottomrule
\end{tabular}
\end{adjustbox}
\label{tab:finetune}
\vspace{-1em}
\end{table}

\subsection{Result Analysis}
\label{sec:analysis}
\begin{figure*}[!htp]
    \vspace{-1.0em}
    \centering
    \includegraphics[width=0.85\textwidth]{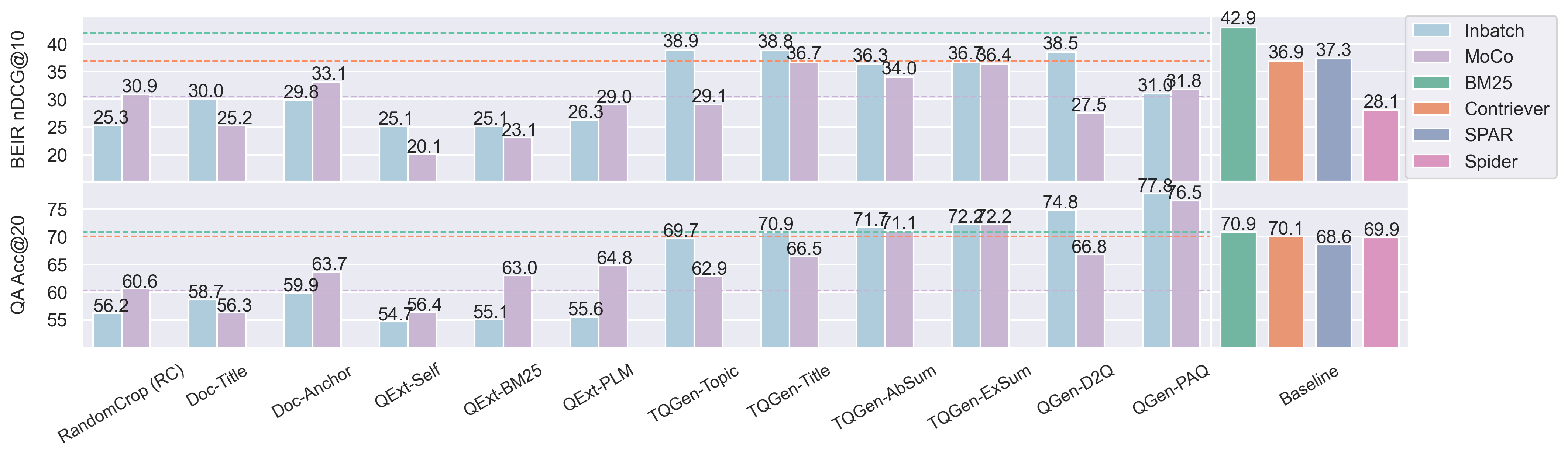}
    \vspace{-1.0em}
    \caption{\augtrv performance with individual augmentation strategies. The upper shows averaged nDCG@10 scores of BEIR benchmark, and the lower shows averaged Recall@20 scores over 6 ODQA datasets. Dashed lines indicate the scores of \bm, \contrv and \textsc{MoCo+RC}.}
    \label{fig:score_wo_RC}
    \vspace{-1.0em}
\end{figure*}

To gain a deeper understanding of how individual augmentation strategies contribute to retrieval tasks and their performance in specific scenarios, we conducted a detailed analysis. We present the results of \augtrv (trained on \textsc{AugQ-Wiki}) and baseline models in Fig~\ref{fig:score_wo_RC}, averaging the scores across 14 BEIR datasets and 6 ODQA datasets. 
% Notably, while BM25 remains a competitive unsupervised baseline and leads among all models, 
We made the following observations:
% 1. MoCo performs better with RC
% 2. InBatch performs similarly with either title or anchor, whereas MoCo favors anchor
% 3. TQGen leads to the best performance, in certain cases they are comparable with QGen settings.
% 4. Among three settings of QExt, PLM is best
% 5. PAQ performs superior well on ODQA, but not generalize well on BEIR, suggesting a domain preference.

\noindent\textbf{1. BEIR is a more comprehensive benchmark for evaluating retrieval models}. In general, the trends observed on BEIR align with those on ODQA. However, ODQA datasets are specifically designed for question-answering using Wikipedia, which introduces certain domain and task biases. For example, \textsc{QGen-PAQ}, which is trained with 65M generated query-document pairs on Wikipedia, excels on ODQA (in-distribution) but fails to generalize well on BEIR. On the other hand, BEIR covers a wider range of domains and topics, making it a more suitable benchmark for evaluating models' generalization ability. Thus, we consider BEIR to be a more indicative benchmark for text retrieval evaluation and focus our discussion on it.
% On the other hand, all ODQA datasets are considered in-distribution since models are trained with Wikipedia data. Since QA queries often require capturing lexical variants and semantic relationships, dense models demonstrate distinct advantages over BM25. 
% This suggests BEIR can be a more indicative benchmark for evaluating text retrieval, and we use it for discussion in the rest of the paper.

\noindent\textbf{2. Among all \augtrv variants, \textsc{TQGen} achieves the highest scores}, significantly outperforming all dense baselines. 
% This a strong evidence indicating that the outputs of language generation tasks, e.g. keyword/title/summary generation, can be directly carried over into training dense retrieval models, without utilizing any annotated queries or questions. 
This finding strongly suggests that the outputs of language generation tasks, such as keyword/title/summary, can be directly utilized for training dense retrieval models. Interestingly, shorter pseudo queries (\textsc{Topic/Title}) perform better on BEIR compared to \textsc{QGen-D2Q}. In contrast, longer ones (\textsc{AbSum/ExSum}) are more effective on ODQA, likely due to their resemblance to questions by including more details. \textsc{ExSum} slightly outperforms \textsc{AbSum}, possibly because it tends to use original text and has fewer hallucinations.

% Previous studies have hypothesized that summaries can be a surrogate for actual queries~\cite{macavaney2017title4query,shen2022lowresource,mass2020weaksupervision}, and our results reveal that 
\noindent\textbf{3. InBatch benefits from queries of higher quality, while MoCo performs well with noisy queries}. The \textsc{RandomCrop} strategy generates noisy queries, including incomplete sentences and non-informative text. However, MoCo is able to achieve good results with such noisy queries, indicating that a momentum encoder can provide robustness against noisy pairs. Conversely, InBatch performs notably better with "cleaner" queries (\textsc{TQGen} and \textsc{QGen}), highlighting the advantages of each architecture.

\begin{figure*}[!ht]
    \centering
    % \vspace{-0.8em}
    \includegraphics[width=0.85\textwidth]{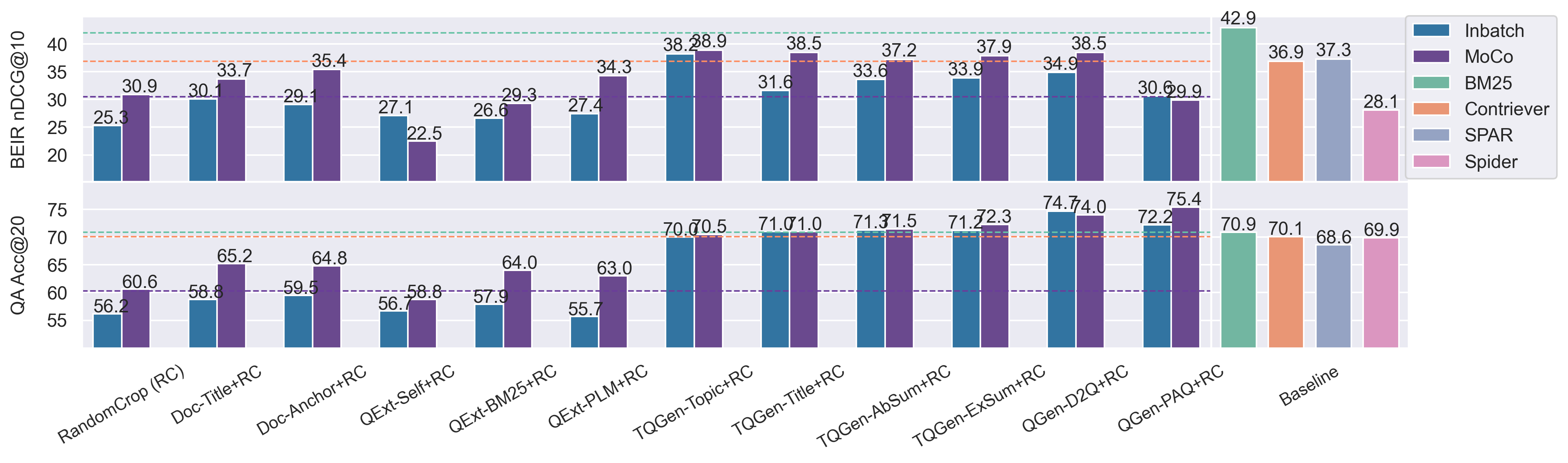}
    \vspace{-1.5em}
    \caption{\augtrv performance with hybrid strategies (50/50 mix of training examples from \textsc{RandomCrop} and another augmentation).}
    \label{fig:score_w_RC}
    \vspace{-1.5em}
\end{figure*}

\noindent\textbf{4. Combining \textsc{RandomCrop} with other high-quality queries benefits MoCo, but not InBatch}. In Figure~\ref{fig:score_w_RC}, we observe that MoCo consistently improves performance by incorporating individual augmentation methods with \textsc{RandomCrop}. This demonstrates MoCo's advantage to leverage multiple strategies and enhance its generalization capability. However, the mixed strategy provides little benefit or worsens the performance for InBatch, which aligns with our previous argument regarding the characteristics of each architecture. We also find that the mixed strategy helps MoCo achieve decent results with QExt methods (\textsc{Doc-Title}/\textsc{Doc-Anchor}/\textsc{QExt-PLM}), though they lag behind \textsc{TQGen} by a significant margin.

% \noindent \textbf{4.} Following the previous finding, we wonder whether combining \textsc{RandomCrop} (noisy but diverse) with other better quality queries can lead to better results. In Figure~\ref{fig:score_w_RC}, we find that \textbf{MoCo obtains consistent gains by mixing individual augmentation methods with \textsc{RandomCrop}}, performing on a par with the best results of InBatch without \textsc{RandomCrop}. This manifests that MoCo can take advantage of multiple strategies, resulting in better generalization ability. Conversely, the mixed strategy causes little help or even a performance drop on InBatch, which echoes our previous argument on the traits of each architecture. 
% We also find that the mixed strategy helps MoCo achieve decent results with \textsc{Doc-Title}, \textsc{Doc-Anchor} and \textsc{QExt-PLM}, but they lag behind  \textsc{TQGen} by large margins.

\subsection{Qualitative Analysis of \textsc{QExt} and \textsc{TQGen}}
\label{sec:qext-qualitative}
Two examples of random spans ranked by two proposed \textsc{QExt} methods are shown in Table~\ref{tab:qext-qualitative}. In the first example, we notice that both \textsc{BM25} and \textsc{QEXT-PLM} are able to rank informative spans to the higher places and place generic spans to the bottom (e.g. ``a few years later'' and ``over the first three''). In the second example, we observe that \textsc{BM25} is more influenced by the low-frequent words (e.g. HS, HO, NaH), whereas \textsc{QEXT-PLM} is more resistant to the noise, placing more informative spans to the top.

Three documents from Pile-CC and the corresponding generated pseudo queries are shown in Table~\ref{tab:qgen-qualitative}. We find that most generated outputs are semantically relevant, in spite of a certain degree of hallucination. In most cases, \textsc{Topic} outputs one or two important phrases and \textsc{Title} outputs one short sentence. Both \textsc{AbSum} and \textsc{ExSum} generate relatively longer sentences as summaries, and \textsc{ExSum} does not necessarily use contents of the original texts. % In the last example, four outputs by T0 are almost the same, indicating that the model ignores the specified prompts.


\section{Related Work}
% \noindent\textbf{Unsupervised Neural Retrieval}
Recent years have seen a flourishing of research works for neural network based information retrieval and question answering. The interested reader may refer to ~\cite{lin2021pretrained,guo2022semantic,zhao2022dense} for a comprehensive overview. Our study, along with a line of recent studies~\cite{contriever,spider}, falls under the category of self-supervised learning using contrastive learning~\cite{shen2022lowresource}, in which a model is trained to maximize the scores of positive pairs and minimize the scores of negative ones. It has demonstrated effective for supervised dense retrieval~\cite{dpr,ance,liu2021dense} and pretraining~\cite{contriever,yu2022coco}. Different from most prior studies, we target at unsupervised models that can be independently applied in retrieval tasks, without any further tuning using annotated data.

% Previous works propose different ways to construct query-document pairs to fit the requirement of contrative learning. \citeauthor{ict} propose inverse cloze task (ICT), using a random sentence as a pseudo query to predict the surrounding context in a batch. REALM~\cite{realm} pretrains a retriver and generator with a pair of a salient span (named entities) and its context. Spider~\cite{spider} proposes to use recurring spans as pseudo queries. The above studies focus on ODQA tasks and their pseudo queries tend to be entity-like, but results of this study and \citeauthor{contriever} show that entity-like queries (e.g. anchor texts) fail to generalize well in a broad range of domains. Some studies propose more generic ideas for training unsupervised models. Specifically, \citeauthor{cpt-paper, costa} use neighboring pieces of text as positive pairs. \citeauthor{contriever} adopt a random cropping strategy to sample two text spans, encouraging the model to learn lexical matching. \citeauthor{spar} use random sentences or real questions as queries and pair them with documents ranked by \bm.

A few research works investigate techniques of data augmentation and domain adaptation for text retrieval and understanding~\cite{tang2022augcse,gpl,iida2022unsupervised}. Query and question generation have been shown as an effective method for augmenting retrieval training data~\cite{beir,doc2query,paq,ma2021zero,gangi-reddy-etal-2022-towards,cho2022query,liang2020embedding}. 
GPL~\cite{gpl} uses cross-encoder to select a good set of synthetic query-document pairs for domain adaptation.
InPars and Promptagator~\cite{inpars,dai2022promptagator} propose to generate questions using large language models in a few-shot manner. LaPraDoR~\cite{laprador} propose to use Dropout-as-Positive-Instance for pretraining retrievers. CERT~\cite{cert} uses positive pairs generated by back-translation. HyDE~\cite{gao2022hyde} uses large language models to augment user queries with generated pseudo documents.
For hyperlinks, \citeauthor{chang2019ICT+WLP+BFS} compare three pretraining tasks for retrieval -- inverse cloze task, body first selection, and wiki link prediction. 
\citeauthor{zhou2022hyperlink,wu2022hyperlink,xie2023webanchor} utilize hyperlinks to construct pseudo query-document pairs. PROP~\cite{ma2021prop,ma2021bprop} uses a representative words prediction task to optimize the semantic distance between a document and a pair of random word sets, estimated by language models. 
Recent studies UPR~\cite{sachan2022improving} and ART~\cite{sachan2022questions} use pretrained language models for reranking and Open-domain QA, using the likelihood of question generation to approximate the relevance between questions and documents. DRAGON~\cite{lin2023dragon} systematically examine supervised training of dense retriever under the framework of data augmentation.

Recently, several studies, such as GTR~\cite{ni2021gtr}, E5~\cite{wang2022e5}, GTE~\cite{li2023gte} and BGE~\cite{bge}, leverage a large amount of paired data to pretrain and fine-tune dense retrievers. They have achieved great performance on BEIR and MTEB benchmarks, but their drawback is that they heavily rely on numerous labeled data. Our work focuses on developing dense retrievers without the reliance on labeled data and this is orthogonal to their motivation. Another line of research (InstructOR~\cite{su2022instructor}, TART~\cite{asai2022tart}, E5Mistral~\cite{wang2023e5mistral}, SFR-Embedding~\cite{sfrembedding}) explores the value of instructions and they demonstrate that embedders can adapt to various tasks with appropriate prompts.
% Recent efforts also consider utilizing pretrained large language models to generate pseudo queries in a few-shot manner~\cite{dai2022promptagator, bonifacio2022inpars} or augment user queries~\cite{gao2022precise}, which shed light on another direction that how to employ more power models for information retrieval. Nevertheless it is also worth considering how to 

\section{Discussion and Conclusion}
In this study, we propose a set of scalable augmentation methods to generate surrogate queries for training dense retrievers without the need for annotated query-document pairs. Our approach achieves great performance on widely used benchmarks (BEIR and six ODQA tasks). These results highlight the effectiveness of extracted and transfer generated query-document pairs for training dense retrievers and prompt us to consider low-cost alternatives in place of expensive human annotations.
For future research, an open question remains regarding the differences between synthetic and real query-document pairs. It would be interesting to explore how various augmentation methods contribute to dense retrieval and investigate the salient span selection for query extraction in more depth. 
% Factors such as span length and filtering methods are worth studying in future work, as they have the potential to impact performance. 
% Besides, we demonstrate that combining multiple augmentation methods leads to further improvements. 
% We are also intrigued by the possibilities of exploring hybrid strategies to enhance retrieval performance and plan to investigate other combinations of augmentation methods in pursuit of this goal.

\appendix

\begin{table*}[!hbt]
\tiny
\caption{Examples of random spans ranked by \bm and a Pretrained Language Model (T5-Small). The numbers in brackets are scores of \bm or T5-Small (negative log likelihood).}
\vspace{-1em}

\begin{adjustbox}{width=1.0\textwidth}
\begin{tabular}[m]{L{6cm} L{6cm} L{6cm}}
    \hline
    \multicolumn{1}{c }{\textbf{Document}} & 
    \multicolumn{1}{c }{\textbf{BM25}} & 
    \multicolumn{1}{c}{\textsc{\textbf{QEXT-PLM}} (T5-Small)} \\ \hline

\multirow{16}{6cm}{ASD can sometimes be diagnosed by age 14 months, although diagnosis becomes increasingly stable over the first three years of life: for example, a one-year-old who meets diagnostic criteria for ASD is less likely than a three-year-old to continue to do so a few years later. In the UK the National Autism Plan for Children recommends at most 30 weeks from first concern to completed diagnosis and assessment, though few cases are handled that quickly in practice. Although the symptoms of autism and ASD begin early in childhood, they are sometimes missed; years later, adults may seek diagnoses to help them or their friends and family understand themselves, to help their employers make adjustments, or in some locations to claim disability living allowances or other benefits.} & {[}14.09{]} diagnosis and assessment, though few cases are handled & {[}79.73{]} adjustments, or in some locations to claim disability \\
 & {[}14.05{]} partly because autistic symptoms overlap with those & {[}85.62{]} completed diagnosis and assessment, though few cases are \\
 & {[}13.55{]} completed diagnosis and assessment, though few cases are & {[}90.46{]} some locations to claim disability living allowances or \\
 & {[}12.06{]} adjustments, or in some locations to claim disability & {[}102.39{]} diagnosis and assessment, though few cases are handled \\
 & {[}11.05{]} some locations to claim disability living allowances or & {[}105.74{]} partly because autistic symptoms overlap with those \\
 & {[}10.48{]} the challenge of obtaining payment can & {[}112.83{]} and assessment, though few cases are handled \\
 & {[}9.78{]} Conversely, the cost of screening & {[}115.09{]} ASD begin early in childhood, \\
 & {[}9.67{]} ASD begin early in childhood, & {[}120.57{]} family understand themselves, to help their \\
 & {[}9.42{]} in some locations to claim disability & {[}127.31{]} Conversely, the cost of screening \\
 & {[}9.38{]} and assessment, though few cases are handled & {[}135.09{]} the challenge of obtaining payment can \\
 & {[}8.45{]} overlap with those of common blindness & {[}139.24{]} overlap with those of common blindness \\
 & {[}8.15{]} family understand themselves, to help their & {[}139.38{]} friends and family understand themselves, \\
 & {[}7.03{]} facial expressions and eye & {[}141.17{]} in some locations to claim disability \\
 & {[}6.77{]} friends and family understand themselves, & {[}147.92{]} a few years later. \\
 & {[}5.56{]} a few years later. & {[}159.65{]} facial expressions and eye \\
 & {[}4.32{]} over the first three & {[}180.59{]} over the first three \\
 \hline

\multirow{16}{6cm}{Reaction with oxygen Upon reacting with oxygen, alkali metals form oxides, peroxides, superoxides and suboxides. However, the first three are more common. The table below shows the types of compounds formed in reaction with oxygen. The compound in brackets represents the minor product of combustion. The alkali metal peroxides are ionic compounds that are unstable in water. The peroxide anion is weakly bound to the cation, and it is hydrolysed, forming stronger covalent bonds. NaO + 2HO → 2NaOH + HO The other oxygen compounds are also unstable in water. 2KO + 2HO → 2KOH + HO + O LiO + HO → 2LiOH Reaction with sulphur With sulphur, they form sulphides and polysulphides. 2Na + 1/8S → NaS + 1/8S → NaS...NaS Because alkali metal sulphides are essentially salts of a weak acid and a strong base, they form basic solutions. S + HO → HS + HO HS + HO → HS + HO Reaction with nitrogen Lithium is the only metal that combines directly with nitrogen at room temperature.} & {[}26.35{]} HS + HO HS + HO → & {[}207.94{]} 2NaCl Alkali metals in liquid ammonia Alkali metals \\
 & {[}25.02{]} NaH + HO → NaOH + H Reaction & {[}227.13{]} they form sulphides and polysulphides. 2Na + 1/8S \\
 & {[}23.74{]} 2NaCl Alkali metals in liquid ammonia Alkali metals & {[}256.10{]} give dilithium acetylide. Na and K can react \\
 & {[}22.44{]} + HO → NaOH + H Reaction & {[}298.67{]} Reaction with sulphur With sulphur, \\
 & {[}20.70{]} Na + xNH → Na + & {[}304.82{]} dilithium acetylide. Na and K \\
 & {[}18.01{]} + 1/3N → LiN LiN & {[}309.33{]} 4NaCl + Ti Reaction with organohalide \\
 & {[}18.00{]} (at 150C) Na + NaCH → & {[}312.97{]} Because alkali metal sulphides are essentially salts of \\
 & {[}17.89{]} + 2HO → 2KOH + HO + & {[}315.61{]} peroxides are ionic compounds that are unstable in \\
 & {[}17.36{]} high temperatures) NaH + HO → & {[}316.52{]} HS + HO HS + HO → \\
 & {[}17.30{]} reaction with oxygen. The compound in brackets & {[}320.61{]} NaH + HO → NaOH + H Reaction \\
 & {[}16.25{]} give dilithium acetylide. Na and K can react & {[}321.21{]} + 2HO → 2KOH + HO + \\
 & {[}15.73{]} they form sulphides and polysulphides. 2Na + 1/8S & {[}327.07{]} give dilithium acetylide. Na \\
 & {[}15.60{]} Reaction with sulphur With sulphur, & {[}336.63{]} Na + e(NH) Due to the presence \\
 & {[}15.39{]} Alkali metals dissolve in liquid ammonia or & {[}337.68{]} the case of Rb and Cs. Na \\
 & {[}15.30{]} Because alkali metal sulphides are essentially salts of & {[}347.56{]} (at 150C) Na + NaCH → \\
 & {[}14.90{]} the case of Rb and Cs. Na & {[}348.45{]} 1/2H (at 150C) Na + NaCH \\
 \hline
    \end{tabular}
    \end{adjustbox}
    \label{tab:qext-qualitative}
\end{table*}

\begin{table*}[hbt!]
\tiny
\caption{Examples of generated pseudo queries by \textsc{TQGen} and \textsc{QGen}.}
\vspace{-1em}

\renewcommand{\arraystretch}{1.25}
    \begin{adjustbox}{width=1.0\textwidth}
\begin{tabular}{lL{6cm}}
\hline
\multicolumn{1}{c}{\textbf{Document}} & 
\multicolumn{1}{c}{\textbf{Output}} \\ \hline
\multirow{5}{8cm}{Topic: reinvent midnight madness\textbackslash{}n\textbackslash{}nAmazon announced a new service at the AWS re:Invent Midnight Madness event. Amazon Sumerian is a solution that aims to make it easier for developers to build virtual reality, augmented reality, and 3D applications. It features a user friendly editor, which can be used to drag and drop 3D objects and characters into scenes.} 
 & \textbf{Topic:} Sumerian \\
 & \textbf{Title:} aws announces sumerian 3d editor \\
 & \textbf{AbSum:} Developing immersive experiences for virtual and augmented reality (VR) and 3D gaming has become an increasing challenge for tech companies. \\
 & \textbf{ExSum:} a new service called amazon sumerian is launched \\
 & \textbf{QGen:} what is amazon sumerian 3d? \\
\hline

\multirow{5}{8cm}{\\The Difference Between Botox and Dermal Fillers\textbackslash{}n\textbackslash{}nWritten by CG Cosmetic on February 19, 2015\textbackslash{}n\textbackslash{}nCG Cosmetic understands the difficulties that come with aging. Everyone has days where they look in the mirror and are concerned with what they see. Whether or not you see deep lines and wrinkles, or fine lines, aging is inevitable. The question then, is what can you do about these signs of aging? Perhaps the most common solution people have heard of is Botox. However, Botox is not the only option. While CG Cosmetic offers Botox procedures, we also offer Dermal Fillers.\textbackslash{}n\textbackslash{}nBotox\textbackslash{}n\textbackslash{}nMost men and women have heard about celebrities using Botox to rejuvenate their skin and appear more youthful, but Botox isn\textbackslash{}u2019t just for celebrities. Botox has provided amazing and effective results that are also safe and convenient for many individuals. Facial lines and wrinkles often occur because of the way your muscles work underneath your skin...\\} 
 & \textbf{Topic:} Botox vs. Dermal Fillers \\
 & \textbf{Title:} Botox is popular for many reasons, including it helping treat aging lines and wrinkles. Dermal fillers, meanwhile, add volume to \\
 & \textbf{AbSum:} Botox relaxes muscles under the skin and helps treat aging lines and wrinkles. Dermal Fillers, meanwhile, add volume to treat sagging skin. CG Cosmetic offers both treatments to help patients maintain a youthful appearance. \\
 & \textbf{ExSum:} Botox relaxes muscles under the skin and helps treat aging lines and wrinkles. Dermal Fillers aims to add volume to facial features such as lips, chin, and jaw line. \\
 & \textbf{QGen:} is botox used for aging? \\

\hline

\multirow{5}{8cm}{\\Cryopreserved (frozen) Donor vials are available to individual Client(s)/Recipient(s), within the United States and Worldwide to achieve Assisted Reproduction.\textbackslash{}nBioGenetics Corporation was established in 1980 to become the first commercial sperm bank in the United States located in New Jersey.\textbackslash{}n\textbackslash{}nBioGenetics respects the current medical standards and ethics set forth by the:\textbackslash{}n\textbackslash{}nAmerican Society for Reproductive Medicine (ASRM)\textbackslash{}nAmerican Society of Andrology (ASA)\textbackslash{}nThe American Association of Tissue Banks (AATB)\textbackslash{}nThe American Urological Association (AUA)\textbackslash{}nand all related scientific associations as well as regulatory government agencies that\textbackslash{}nmay impact Reproductive Cell and Tissue Banking...\\}
 & \textbf{Topic:} Cell banking \\
 & \textbf{Title:} BioGenetics is a Commercial Sperm Bank and Tissue Bank \\
 & \textbf{AbSum:} A commercial sperm bank for patients seeking fertility treatment and assisted reproduction \\
 & \textbf{ExSum:} BioGenetics is a Commercial Sperm Bank and Tissue Bank providing services to all Assisted Reproductive Technologies. \\
 & \textbf{QGen:} where is biogenetics located? \\

 \hline

    \end{tabular}
    \end{adjustbox}
    \label{tab:qgen-qualitative}
\vspace{-1em}
\end{table*}

\bibliographystyle{ACM-Reference-Format}
\bibliography{sample-base,anthology,custom}

\end{document}